# A New Unified Deep Learning Approach with Decomposition-Reconstruction-Ensemble Framework for Time Series Forecasting

Guowei Zhang[a,b], Tao Ren[a,b], Yifan Yang[c*]

[a] Academy of Mathematics and Systems Science, Chinese Academy of Sciences, Beijing, 100190, China

[b] School of Mathematical Sciences, University of Chinese Academy of Sciences, Beijing, 100049, China

[c] School of Management, Xi'an Jiaotong University, Xi'an, 710049, China

[*] Corresponding author. E-mail address: jixieyyf2014@stu.xjtu.edu.cn (Y.F. Yang).

**Abstract:** A new variational mode decomposition (VMD) based deep learning approach is proposed in this paper for time series forecasting problem. Firstly, VMD is adopted to decompose the original time series into several sub-signals. Then, a convolutional neural network (CNN) is applied to learn the reconstruction patterns on the decomposed sub-signals to obtain several reconstructed sub-signals. Finally, a long short term memory (LSTM) network is employed to forecast the time series with the decomposed sub-signals and the reconstructed sub-signals as inputs. The proposed VMD-CNN-LSTM approach is originated from the decomposition-reconstruction-ensemble framework, and innovated by embedding the reconstruction, single forecasting, and ensemble steps in a unified deep learning approach. To verify the forecasting performance of the proposed approach, four typical time series datasets are introduced for empirical analysis. The empirical results demonstrate that the proposed approach outperforms consistently the benchmark approaches in terms of forecasting accuracy, and also indicate that the reconstructed sub-signals obtained by CNN is of importance for further improving the forecasting performance.

**1. Introduction**

Time series forecasting is an important and challenging research topic that has been investigated for several years [1,2]. It is particularly useful and has a wide application in many fields, such as financial time series forecasting [3–6], crude oil price forecasting [7–13], wind speed forecasting [14–17], traffic flow forecasting [18–20], energy consumption forecasting [21–23] and so on. Thus, more and more effort has been devoted over the past decades to the development and improvement of time series forecasting approaches.

Many econometric and statistical models have been used for time series forecasting problems, such as autoregressive integrated moving average (ARIMA) [24–28], co-integration models [21,29], generalized autoregressive conditional heteroscedasticity (GARCH) [11,27,30,31], vector auto-regression (VAR) [21,32], error correction models (ECM) [21,33,34], and linear regression (LR). Although these models achieve good performance in several time series forecasting problems, they gain poor performance for time series with complex nonlinear patterns. Thus, to capture the nonlinearity and complexity of the time series datasets, several nonlinear and more complex models are introduced for time series forecasting problems, such as support vector regression (SVR) [3–5,26,35], random forest regression (RFR) [22], and extreme learning machine (ELM) [14,36–39]. These models provide flexible nonlinear modeling capability and achieve better performance in several complex time series forecasting problems. In recent years, deep learning models obtained a booming development for achieving a state-of-the-art accuracy on several challenging problems [40,41], such as convolutional neural network (CNN) for image recognition, recurrent neural network (RNN) or long short term memory network (LSTM) for speech recognition and natural language processing, auto-encoder for feature extraction. Due to the good performance of deep learning models on forecasting problems, many researchers bring deep learning models into the field of time series forecasting [14,15,18,19,42–44]. Using hybrid model or combining several models has become a common practice to improve the forecasting performance. Therefore, to further improve the performance of deep learning models, several researchers combined different deep learning models into a hybrid model for time series forecasting, and it obtained better performance than single deep learning models [45–47]. While these studies always employ different deep learning models separately, how to combine different deep learning models into a unified model deserves more research attentions.

The decomposition-ensemble framework proposed by Yu et al. [12] is a promising hybrid model for complex time series forecasting, which applies different models for different tasks to obtain a better performance, and it has become more and more popular in recent years. The main idea of decomposition-ensemble framework is decomposing the complex time series into several sub-signals and forecasting the sub-signals respectively instead of forecasting the original time series. Three different steps are contained in the decomposition-ensemble framework, namely decomposition step, single forecasting step, and ensemble step. The decomposition step decomposes the original time series into several sub-signals, the single forecasting step forecasts each sub-signals to obtain sub-forecasting results, and the ensemble step combines the sub-forecasting results to obtain the final forecasting results. Since the sub-signals always are easier to be forecasted, the decomposition-ensemble framework always obtain a better performance. The decomposition-ensemble framework provides a novel forecasting paradigm, and has been widely and successfully used in several time series forecasting problems, such as crude oil price forecasting [7–10,13], exchange rate forecasting [48], wind speed forecasting [14–16], and other time series forecasting [49–51].

Recently, a decomposition-reconstruction-ensemble framework has been developed which extends the decomposition-ensemble framework by adding a reconstruction step before single forecasting step and obtained a better performance [52]. With the reconstruction step, the decomposed sub-signals can be combined into different reconstructed sub-signals, which tend to be easier for modeling and forecasting than the decomposed sub-signals, thus a better performance can be obtained by the decomposition-reconstruction-ensemble framework. The decomposition-reconstruction-ensemble framework provide an extension and scalability to the decomposition-ensemble framework, and promising hybrid models can be built based on the framework for time series forecasting problems. In addition, it can also provide a solution to combine different deep learning models by employing them in different steps. Moreover, since neural network models can unify different deep learning models, the different steps in the decomposition-reconstruction-ensemble framework can be embedded into a unified deep learning approach, which innovate the previous studies which perform each step of decomposition-reconstruction-ensemble framework separately. Since the unified deep learning approach can be trained end-to-end, it can provide better performance than the model which performs separately.

Therefore, we propose a new variational mode decomposition based deep learning approach, which is originated from the decomposition-reconstruction-ensemble framework. The main innovation of the proposed approach is that the reconstruction, single forecasting, and ensemble are embedded in a unified deep learning approach. In the unified deep learning approach, CNN is employed for reconstruction step and LSTM is employed for single forecasting and ensemble steps. To the best of our knowledge, this is the first study that directly applies deep learning on the obtained sub-signals of the time series and embeds the construction, single forecasting and ensemble steps in a unified deep learning approach.

In this paper, we will investigate the performance of the proposed VMD-CNN-LSTM approach: (1) showing how to construct the proposed approach; (2) validating the effectiveness of the proposed approach by comparing with benchmark models; (3) exploring the impact of the reconstruction step for the performance of the proposed approach.

The remainder of this paper is organized as follows. The related methodologies are introduced in Section 2. Section 3 gives the proposed VMD-CNN-LSTM approach for time series forecasting. The empirical analysis is conducted in Section 4. Finally, Section 5 draws the conclusions.

## 2. Related Methodologies

### 2.1 Variational mode decomposition

Variational mode decomposition (VMD) is proposed by Dragomiretskiy and Zosso in 2014 [53], which is an entirely non-recursive signal processing technique, and can adaptively decompose a signal into several band-limited sub-signals (i.e., modes). It is assumed that each mode ($\mu_k$) of the original signal mostly centered around a center pulsation ($\omega_k$) and is determined during the time of decomposition process. The process of assessing the bandwidth of each mode is as follows: the associated analytic signal of each mode is calculated by Hilbert transform, so that a unilateral frequency spectrum is obtained; the frequency spectrum of each mode is shifted to baseband by mixing with an exponential tuned to the respective estimated center frequency; and the bandwidth of each mode is estimated through $H^1$ Gaussian smoothness of the demodulated signal. Thus, VMD can be implemented by solving a resulting constrained variational optimization problem as follows:

$$\min_{\{\mu_k\},\{\omega_k\}} \sum_{k=1}^{K} \left\| \partial_t \left[ \left( \delta(t) + \frac{j}{\pi t} \right) \otimes \mu_k(t) \right] e^{-j\omega_k t} \right\|_2^2,$$
$$s.t. \sum_{k=1}^{K} \mu_k(t) = f(t) \tag{1}$$

where $f(t)$ is the original signal, $\mu_k(t)$ is the $k$th component of $f(t)$, $\omega_k$ is the center pulsation of $\mu_k$, $\{\mu_k\} = \{\mu_1, \mu_2, \ldots, \mu_K\}$ represents the set of all modes, $\{\omega_k\} = \{\omega_1, \omega_2, \ldots, \omega_K\}$ represents the set of all center pulsations, $K$ is the number of modes, $\delta(t)$ is the Dirac distribution, $\otimes$ represents convolution operator, $t$ is time script.

The above constrained problem can be converted into an unconstrained problem by considering a quadratic penalty term and Lagrangian multipliers. The augmented Lagrangian is given as follows:

$$L(\{\mu_k\},\{\omega_k\},\lambda)$$
$$= \alpha \sum_{k=1}^{K} \left\| \partial_t \left[ \left( \delta(t) + \frac{j}{\pi t} \right) \otimes \mu_k(t) \right] e^{-j\omega_k t} \right\|_2^2,$$
$$+ \left\| f(t) - \sum_{k=1}^{K} \mu_k(t) \right\|_2^2 + \left\langle \lambda(t), f(t) - \sum_{k} \mu_k(t) \right\rangle \tag{2}$$

where $\alpha$ represents the balancing parameter of the data-fidelity constraint and $\lambda$ represents the Lagrangian multipliers.

Then, the solution of original optimization problem can be found as the saddle point of the augmented Lagrangian through using the alternate direction method of multipliers (ADMM).

Consequently, the solutions of $\mu_k$ and $\omega_k$ are expressed as follows:

$$\hat{\mu}_k^{n+1} = \frac{\hat{f}(\omega) - \sum_{i \neq k} \hat{\mu}_i(\omega) + \frac{\hat{\lambda}(\omega)}{2}}{1 + 2\alpha(\omega - \omega_k)^2}, \tag{3}$$

$$\hat{\omega}_k^{n+1} = \frac{\int_0^\infty \omega |\hat{\mu}_k(\omega)|^2 d\omega}{\int_0^\infty |\hat{\mu}_k(\omega)|^2 d\omega}, \tag{4}$$

where $n$ is the number of iterations, $\hat{f}(\omega)$, $\hat{\mu}_i(\omega)$, $\hat{\lambda}(\omega)$ and $\hat{\mu}_k^{n+1}$ are the Fourier transforms of $f(t)$, $\mu_i(t)$, $\lambda(t)$ and $\mu_k^{n+1}(t)$, respectively.

**2.2 Convolutional neural network**

Convolutional neural network (CNN) is a special case of artificial neural network, which proposed by LeCun Yann in 1998 [54], and becomes especially popular since the proposition and success of AlexNet, which uses convolutional neural network to classify the high-resolution images in the LSVRC-2010 ImageNet [55]. Unlike fully connected neural networks like multi-layer perceptron, CNN adopts a structure of local link which shares the common weights in different locality. The specific structure of CNN can greatly reduce the number of parameters and the computational complexity, which avoids CNN with more layers from over-fitting. Moreover, CNN can extract features of multiple scales at different levels through multi-layer and multi-channel convolution and pooling (sub-sampling) operations. This provides CNN with the property of translation invariance, which is of great significance in tasks such as image processing, thus CNN achieves great success in the field of computer vision.

The typical structure of CNN is shown in **Fig. 1**. The convolution layer and pooling layer are the main structures in a typical CNN, and they are always structured successively. The convolution layer applies convolution operation to the input data of this layer and passes the convolution results to the next layer. The pooling layer applies pooling operation to the input data of this layer and passes the pooling results to next layer. Through pooling layer, the number of parameters and the spatial size of the representation are reduced. In the last pooling layer, the data becomes a one-dimensional vector and is connected to a fully connected layer.

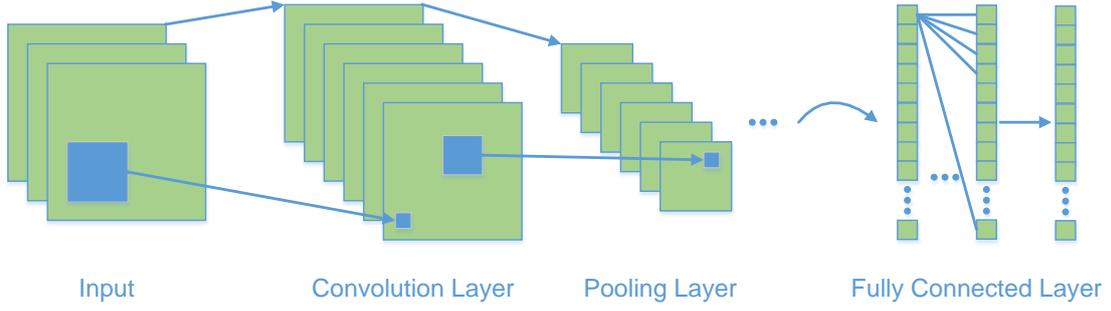

**Fig. 1** The typical structure of CNN

### 2.3 Long short term memory

Long short term memory (LSTM) network was proposed by Sepp Hochreiter and Jürgen Schmidhuber in 1997 [56]. LSTM is a special kind of recurrent neural network (RNN), which is used for processing the sequence data, such as speech, video, and natural language. It is especially suitable for time series problems for the capability of learning arbitrary long-term dependencies in the input data. A common LSTM network consists of input layer, hidden layer and output layer, and the hidden layer is composed of a forget gate, an input gate, an output gate, and a cell. The cell remembers values over arbitrary time intervals and the three gates regulate the flow of information into and out of the cell. The structure of hidden layer of LSTM is shown in **Fig. 2**.

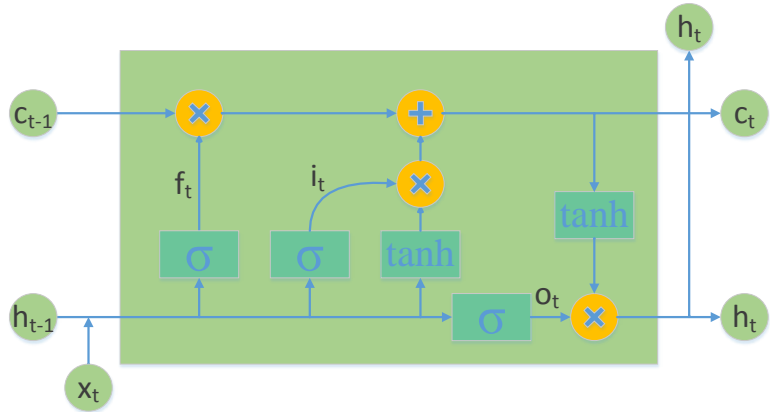

**Fig. 2** The structure of hidden layer of LSTM

The forget rate, input rate, output rate, cell state value, and output of this hidden layer are computed as follows:

$$f_t = \sigma(w_f [h_{t-1}, x_t] + b_f), \tag{5}$$

$$i_t = \sigma(w_i [h_{t-1}, x_t] + b_i), \tag{6}$$

$$o_t = \sigma(w_o [h_{t-1}, x_t] + b_o), \tag{7}$$

$$c_t = f_t c_{t-1} + i_t \tanh(w_c [h_{t-1}, x_t] + b_C), \tag{8}$$

$$h_t = o_t \tanh(c_t), \tag{9}$$

where $f_t$, $i_t$, $o_t$ are forget rate, input rate and forget rate, respectively, $w_f$, $w_i$, $w_o$, $w_c$ are the weights of forget gate, input gate, output gate and cell state layer, respectively, $b_f$, $b_i$, $b_o$, $b_C$ are the bias of forget gate, input gate, output gate and cell state layer, respectively, $x_t$ is the input of this hidden layer, $h_{t-1}$ is the output of last hidden layer, $h_t$ is the output of this hidden layer, $c_{t-1}$ is the cell state value of late hidden layer, $c_t$ is the cell state value of this hidden layer, $\sigma(\cdot)$ is the sigmoid activation function, $\tanh(\cdot)$ is the tanh activation function.

### 3 The proposed VMD-CNN-LSTM approach

The proposed VMD-CNN-LSTM approach is based on the idea of decomposition-reconstruction-ensemble framework, which show a good performance in time series forecasting problem. There are four main steps in the decomposition-reconstruction-ensemble framework: (1) the original time series is decomposed into several sub-signals (decomposition), (2) the sub-signals are combined into several reconstructed sub-signals (reconstruction), (3) some forecasting models are employed to forecast the reconstructed sub-signals respectively (single forecasting), (4) the forecasting results of each reconstructed sub-signals are combined to obtain the final forecasting results (ensemble). In the proposed VMD-CNN-LSTM approach, we innovate the decomposition-reconstruction-ensemble framework by embedding the reconstruction, single forecasting, and ensemble steps into a unified deep learning approach.

The flowchart of the proposed VMD-CNN-LSTM approach is shown in **Fig. 3**. As illustrated in **Fig. 3**, the proposed VMD-CNN-LSTM approach mainly consists of four steps:

(1) VMD is adopted to decompose the original time series into $K$ sub-signals;

(2) CNN is applied to learn reconstruction patterns on the decomposed sub-signals to

obtain the reconstructed sub-signals (the CNN kernel size is $K$ by $L$, where $L$ is the input sequence length, and a CNN kernel can be regarded as a reconstruction weights for the decomposed sub-signals);

(3) LSTM is employed to forecast the decomposed sub-signals and the reconstructed sub-signals (the decomposed sub-signals and the reconstructed sub-signals can be regarded as a multi-dimensional representation of the original time series);

(4) A fully connected layer is used to combine the forecasting results of LSTM to obtain the final forecasting results.

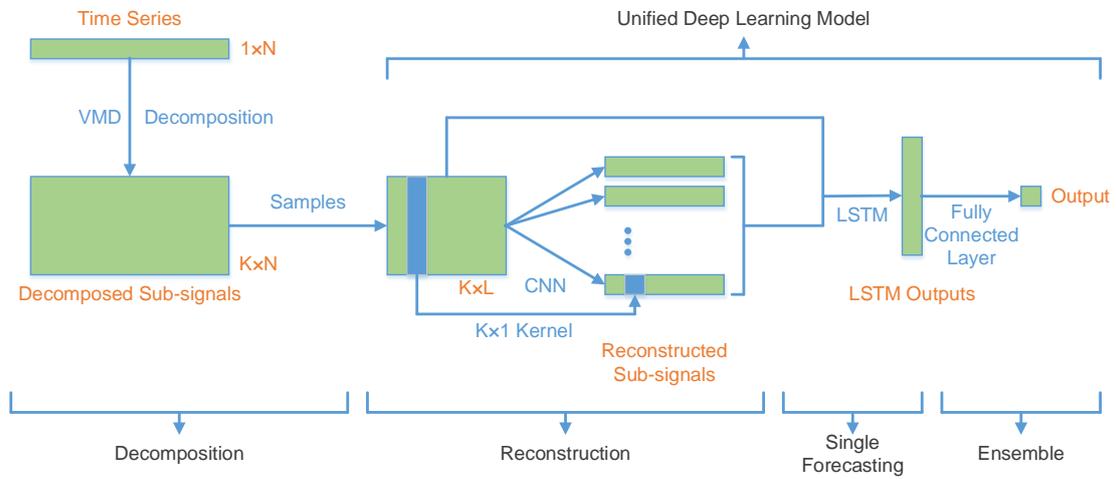

**Fig. 3** The flowchart of the proposed VMD-CNN-LSTM approach

## 4. Empirical analysis

In this section, four time series datasets are introduced to evaluate the effectiveness of the proposed VMD-CNN-LSTM approach. Section 4.1 describes the collected datasets and introduces the evaluation criteria, and Section 4.2 gives the empirical results.

### 4.1 Data description and evaluation criteria

To evaluate the performance of the proposed VMD-CNN-LSTM approach, we choose two typical time series for empirical analysis, i.e., crude oil price time series and wind speed time series. Two crude oil price datasets are collected from the U.S. Energy Information Administration (https://www.eia.gov/), including daily crude oil price of West Texas Intermediate (WTI) crude oil spot price and Brent crude oil spot price. The daily crude oil price of WTI is range from January 3, 2011 to February 28, 2018, excluding weekends and holidays (2023 samples in total), and the daily crude oil price of Brent is range from January 4, 2011 to February 31, 2018, excluding weekends and holidays (2062 samples in total). The wind speed time series datasets are collected from a wind farm in Inner Mongolia, China and the step of the wind speed datasets is 15 min. Two different datasets are selected: the first dataset is range from April 4, 2015 to April 19 (1440 samples in total), and the second dataset is range from July 18, 2015 to August 1, 2015 (1440 samples in total). The descriptions and statistical information of the collected datasets are shown in **Table 1** and **Table 2**, respectively. The original data of the collected datasets are shown in **Fig. 4**. All the datasets are divided into two periods: in-sample period with 80% of all the samples and out-of-sample period with 20% of all the samples.

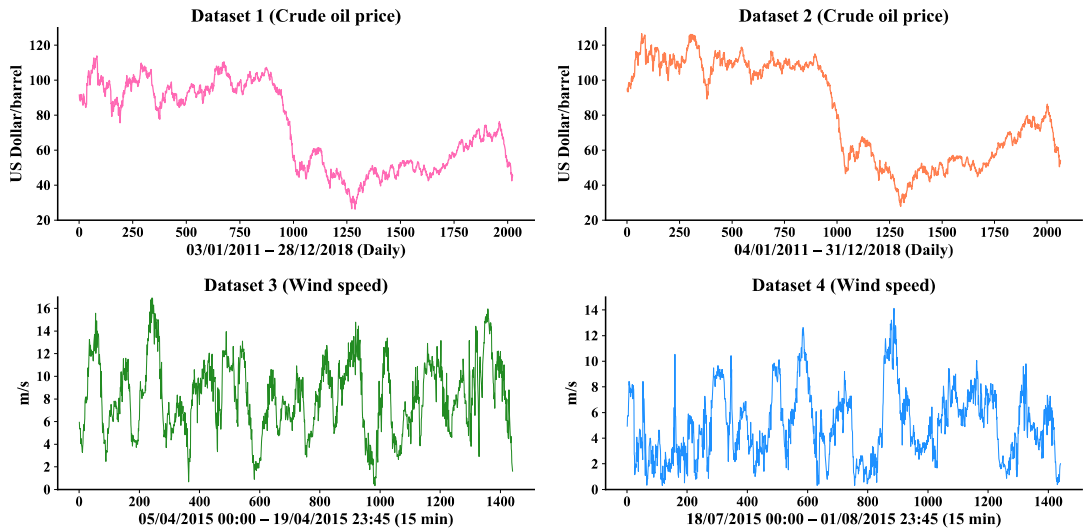

**Fig. 4** The original data of the four selected datasets

**Table 1** The descriptions of the four selected datasets

| Datasets | Time series Type | Time interval | # of samples |
| --- | --- | --- | --- |
| Dataset 1 | Crude oil price (WTI) | 03/01/2011 – 28/12/2018 (Daily) | 2023 |
| Dataset 2 | Crude oil price (Brent) | 03/01/2011 – 28/12/2018 (Daily) | 2062 |
| Dataset 3 | Wind speed | 05/04/2015 00:00 – 19/04/2015 23:45 (15 min) | 1440 |
| Dataset 4 | Wind speed | 18/07/2015 00:00 – 01/08/2015 23:45 (15 min) | 1440 |

**Table 2** The statistical information of the four selected datasets

| Datasets | Minimum | Maximum | Mean | Std. [a] | Skewness | Kurtosis |
|---|---|---|---|---|---|---|
| Dataset 1 | 26.21 | 113.93 | 73.49 | 23.57 | -0.0619 | -1.5324 |
| Dataset 2 | 27.88 | 126.65 | 81.28 | 27.54 | -0.0956 | -1.5939 |
| Dataset 3 | 0.34 | 16.92 | 8.20 | 3.18 | 0.0819 | -0.5019 |
| Dataset 4 | 0.31 | 14.12 | 5.01 | 2.71 | 0.4903 | -0.2961 |

Note: Std. [a] refers to the standard deviation.

In addition, to evaluate the forecasting accuracy of the forecasting approach, we choose mean absolute error (*MAE*), root mean squared error (*RMSE*), and mean absolute percentage error (*MAPE*) as the evaluation criteria in this paper. *MAE*, *RMSE* and *MAPE* are defined as follows:

$$MAE = \sum_{t=1}^{T} |y_t - \hat{y}_t|, \tag{10}$$

$$RMSE = \sqrt{\frac{1}{T} \sum_{t=1}^{T} (y_t - \hat{y}_t)^2}, \tag{11}$$

$$MAPE = \frac{1}{T} \sum_{t=1}^{T} \left| \frac{y_t - \hat{y}_t}{y_t} \right| \times 100\%, \tag{12}$$

where $\hat{y}_t$ is the forecasting value of $y_t$ and $T$ is number of samples.

### 4.2 Empirical results

To validate the performance of the proposed VMD-CNN-LSTM approach, three individual models and three integrated approaches are introduced as benchmark approaches. The individual models consist of RFR, SVR, and LSTM. The integrated approaches consist of VMD-RFR, VMD-SVR, and VMD-LSTM. VMD-RFR, and VMD-SVR are decomposition-ensemble models with VMD for decomposition step, RFR and SVR for single forecasting step respectively, and LR for ensemble step. VMD-LSTM is also a decomposition-ensemble model with VMD for decomposition step and LSTM for single forecasting and ensemble steps. VMD-LSTM can also be regarded a special case of the proposed VMD-CNN-LSTM, which drops the reconstruction structure (without CNN kernels). All the above models are implemented in Python: VMD is implemented with the python package vmdpy, RFR and SVR are implemented with the python package scikit-learn, and CNN and LSTM are implemented with the python package PyTorch.

The kernel function of SVR is the radial basis function (RBF) kernel. the optimal values of hyper-parameters of RFR and SVR are determined by grid search with k-fold cross validation (4 folds are split in this paper). The original time series dataset is decomposed into 4 modes by VMD. The detail of implementation of the deep learning models are as follows: (1) the loss function is mean squared error (MSE), (2) the optimizer is Adam [57], (3) the batch size is 128, (4) the number of training epochs is 2000, (5) the initial learning rate is 0.001, (6) the learning rate is scheduled by cosine annealing with warm restart (restart every 200 epochs in this paper) [58], (7) the activation of CNN is ReLU, and (8) the data of each datasets are normalized between 0 and 1 by min-max normalization. For VMD-LSTM and VMD-CNN-LSTM models, the input sequence length is 12, thus each sample is a 12 by 4 matrix.

For each deep learning models, we choose the number of kernels of CNN ($n_k$) from {1, 3, 5, 7}, the number of hidden layer nodes of LSTM ($n_h$) from {6, 8, 10, 12}, and the number of layers of LSTM ($n_l$) from {1, 2, 3}. By performing grid search, we obtain the optimal values of the hyper-parameters for each deep learning model, which is shown in **Table 3**. In addition, the convergence curve of VMD-LSTM, and VMD-CNN-LSTM with the optimal values of the hyper-parameters are shown in **Fig. 5**.

**Table 3** Parameters setting of the deep learning models

| Dataset | LSTM [a] | | VMD-LSTM [a] | | VMD-CNN-LSTM [b] | | |
|---|---|---|---|---|---|---|---|
| | $n_h$ | $n_l$ | $n_h$ | $n_l$ | $n_k$ | $n_h$ | $n_l$ |
| Dataset 1 | 10 | 2 | 12 | 3 | 5 | 12 | 2 |
| Dataset 2 | 8 | 2 | 10 | 2 | 7 | 10 | 2 |
| Dataset 3 | 6 | 1 | 12 | 2 | 3 | 12 | 2 |
| Dataset 4 | 10 | 1 | 10 | 2 | 1 | 12 | 3 |

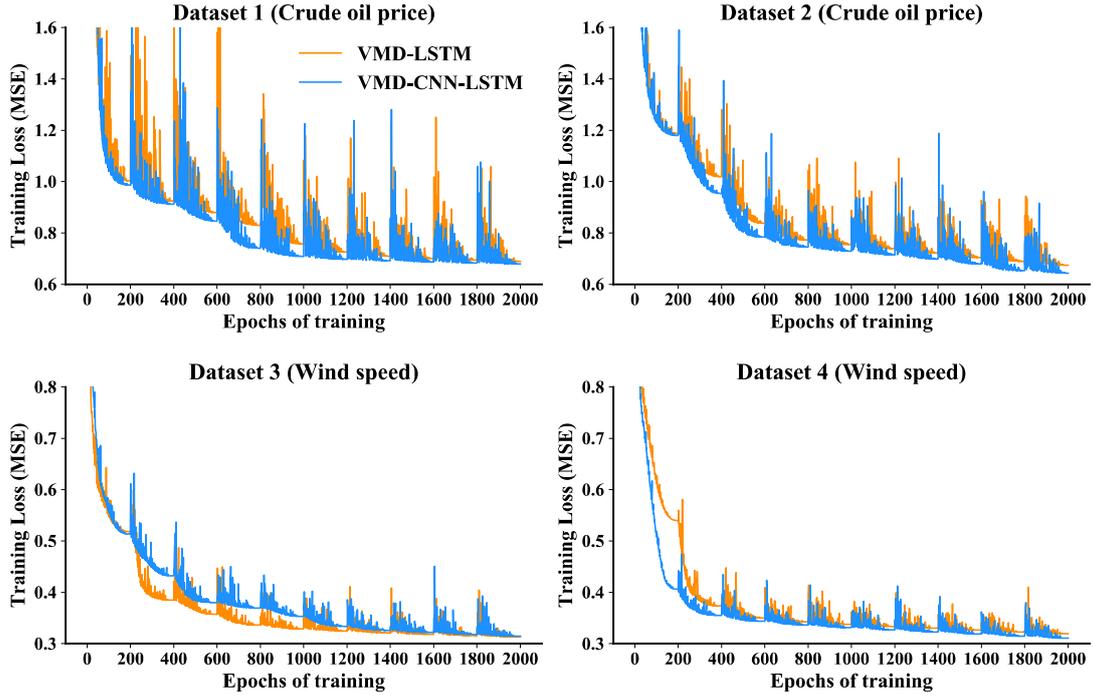

**Fig. 5** Convergence curve of VMD-LSTM and VMD-CNN-LSTM

The proposed approach and the six benchmark approaches are employed to forecast each time series dataset with the optimal values of the hyper-parameters. The in-sample forecasting results and out-of-sample forecasting results of two crude oil price datasets (Dataset 1 and Dataset 2) are shown in **Table 4** and **Table 5**, respectively. The in-sample forecasting results and out-of-sample forecasting results of two wind speed datasets (Dataset 3 and Dataset 4) are shown in **Table 6** and **Table 7**, respectively.

**Table 4** Forecasting results of Dataset 1

| Model | In-sample | | | Out-of-sample | | |
|---|---|---|---|---|---|---|
| | *RMSE* | *MAE* | *MAPE* | *RMSE* | *MAE* | *MAPE* |
| RFR | 1.3151 | 0.9767 | 1.42% | 2.8123 | 2.1683 | 3.39% |
| SVR | 1.4336 | 0.9476 | 1.40% | 1.7985 | 1.3857 | 2.27% |
| LSTM | 1.4083 | 1.0520 | 1.55% | 1.0466 | 0.8054 | 1.36% |
| VMD-RFR | 0.9443 | 0.7082 | 1.03% | 0.9831 | 0.8065 | 1.35% |
| VMD-SVR | 0.9282 | 0.6966 | 1.02% | 0.8686 | 0.6957 | 1.16% |
| VMD-LSTM | 0.6893 | 0.5226 | 0.77% | 0.5199 | 0.3961 | 0.68% |
| VMD-CNN-LSTM | **0.6789** | **0.5180** | **0.76%** | **0.5162** | **0.3933** | **0.67%** |

**Table 5** Forecasting results of Dataset 2

| Model | In-sample | | | Out-of-sample | | |
|---|---|---|---|---|---|---|
| | *RMSE* | *MAE* | *MAPE* | *RMSE* | *MAE* | *MAPE* |
| RFR | 1.3549 | 0.9886 | 1.31% | 3.3623 | 2.4648 | 3.51% |
| SVR | 1.1965 | 0.7803 | 1.02% | 2.0282 | 1.5238 | 2.24% |
| LSTM | 1.4134 | 1.0377 | 1.37% | 1.1729 | 0.8917 | 1.37% |
| VMD-RFR | 0.9560 | 0.7173 | 0.94% | 1.1412 | 0.9423 | 1.42% |
| VMD-SVR | 0.9442 | 0.7058 | 0.93% | 1.0889 | 0.8732 | 1.30% |
| VMD-LSTM | 0.6743 | 0.5096 | 0.67% | 0.5389 | 0.4209 | 0.65% |
| VMD-CNN-LSTM | **0.6437** | **0.4869** | **0.64%** | **0.5323** | **0.4190** | **0.64%** |

**Table 6** Forecasting results of Dataset 3

| Model | In-sample | | | Out-of-sample | | |
|---|---|---|---|---|---|---|
| | *RMSE* | *MAE* | *MAPE* | *RMSE* | *MAE* | *MAPE* |
| RFR | 0.8458 | 0.6300 | 9.76% | 1.3283 | 0.9411 | 11.59% |
| SVR | 0.9272 | 0.6829 | 10.43% | 1.3142 | 0.9327 | 11.52% |
| LSTM | 0.9415 | 0.7085 | 11.00% | 1.3231 | 0.9167 | 11.17% |
| VMD-RFR | 0.5606 | 0.4210 | 6.48% | 0.9401 | 0.6833 | 8.24% |
| VMD-SVR | 0.5540 | 0.4177 | 6.44% | 0.7801 | 0.5524 | 6.82% |
| VMD-LSTM | **0.3131** | 0.2395 | 3.76% | 0.5181 | 0.3634 | 4.62% |
| VMD-CNN-LSTM | 0.3145 | **0.2379** | **3.70%** | **0.4699** | **0.3455** | **4.36%** |

**Table 7** Forecasting results of Dataset 4

| Model | In-sample | | | Out-of-sample | | |
|---|---|---|---|---|---|---|
| | *RMSE* | *MAE* | *MAPE* | *RMSE* | *MAE* | *MAPE* |
| RFR | 0.8011 | 0.5811 | 19.21% | 0.9011 | 0.6519 | 17.42% |
| SVR | 0.8548 | 0.5900 | 19.26% | 0.9330 | 0.6467 | 16.73% |
| LSTM | 0.9038 | 0.6555 | 21.94% | 0.8863 | 0.6439 | 17.33% |
| VMD-RFR | 0.5401 | 0.3972 | 12.98% | 0.5705 | 0.4038 | 10.39% |
| VMD-SVR | 0.5233 | 0.3833 | 12.46% | 0.5291 | 0.3753 | 9.67% |
| VMD-LSTM | 0.3194 | 0.2469 | 8.00% | 0.3376 | 0.2486 | 6.60% |
| VMD-CNN-LSTM | **0.3105** | **0.2409** | **7.82%** | **0.3310** | **0.2432** | **6.38%** |

From the **Table 4-7**, it can be clearly found that the proposed VMD-CNN-LSTM approach outperforms consistently the benchmark models for the four selected datasets in terms of *RMSE*, *MAE*, and *MAPE* (both in-sample forecasting performance and out-of-sample forecasting performance). This suggests that the proposed VMD-CNN-LSTM approach is an effective and promising approach for time series forecasting problems. In addition, we also notice that the forecasting performances of VMD-LSTM and VMD-CNN-LSTM are very close, and significantly better than the other benchmark approaches. This demonstrates that embedding the single forecasting step and ensemble step into a unified deep learning approach is effective to improve the forecasting performance. Comparing the forecasting performance of VMD-CNN-LSTM with VMD-LSTM, it can be found that VMD-CNN-VMD is always better than VMD-LSTM, which demonstrates that the reconstruction step implemented by CNN is helpful to further improve the performance.

By comparing the single models with their corresponding integrated approaches (SVR vs VMD-SVR, RFR vs VMD-RFR, LSTM vs VMD-LSTM), we found that the integrated approaches significantly show better performance than single models, which demonstrates that integrated approaches are a better choice for time series forecasting. By comparing the LSTM based approaches with other approaches (LSTM vs SVR and RFR, VMD-LSTM vs VDM-SVR and VMD-RFR), we found that the LSTM based approaches always show better out-of-sample forecasting performance than other approaches, which demonstrates that LSTM is more suitable for time series forecasting.

In addition, to evaluate forecasting performance of different approaches from a statistical perspective, the Diebold-Mariano (DM) statistic was applied to test the statistical significance of all approaches [59]. The DM statistic was used to test the null hypothesis of equality of expected forecast accuracy against the alternative of different forecasting abilities across approaches. MSE was used as the loss function and the null hypothesis of the DM test was that the out-of-sample MSE of the tested approach is not smaller than that of the benchmark approach. The DM test results are shown in **Table 8-11**. The results suggest that the VMD-CNN-LSTM and VMD-LSTM approaches significantly outperforms other benchmark approaches and the proposed VMD-CNN-LSTM approach is better than VMD-LSTM approach. In addition, the integrated approaches are significantly better than the single models.

**Table 8** DM test results of Dataset 1

| Tested models | Reference model | | | | | |
|---|---|---|---|---|---|---|
| | VMD-LSTM | VMD-SVR | VMD-RFR | LSTM | SVR | RFR |
| VMD-CNN-LSTM | -1.0559 (0.2916) | -11.2610 (0.0000) | -13.1309 (0.0000) | -8.8399 (0.0000) | -11.8110 (0.0000) | -14.0023 (0.0000) |
| VMD-LSTM | | -11.4577 (0.0000) | -13.3321 (0.0000) | -8.8051 (0.0000) | -11.8007 (0.0000) | -14.0039 (0.0000) |
| VMD-SVR | | | -12.2109 (0.0000) | -3.3421 (0.0009) | -10.9567 (0.0000) | -13.6910 (0.0000) |
| VMD-RFR | | | | -1.0816 (0.2801) | -10.5710 (0.0000) | -13.5179 (0.0000) |
| LSTM | | | | | -10.2757 (0.0000) | -13.2213 (0.0000) |
| SVR | | | | | | -6.4088 (0.0000) |

**Table 9** DM test results of Dataset 2

| Tested models | Reference model | | | | | |
|---|---|---|---|---|---|---|
| | VMD-LSTM | VMD-SVR | VMD-RFR | LSTM | SVR | RFR |
| VMD-CNN-LSTM | -0.5149 (0.6067) | -11.2039 (0.0000) | -12.9768 (0.0000) | -7.7163 (0.0000) | -11.0162 (0.0000) | -14.3528 (0.0000) |
| VMD-LSTM | | -11.4078 (0.0000) | -13.1408 (0.0000) | -7.5108 (0.0000) | -11.0093 (0.0000) | -14.3500 (0.0000) |
| VMD-SVR | | | -12.3327 (0.0000) | -1.7696 (0.0775) | -10.5597 (0.0000) | -13.9480 (0.0000) |
| VMD-RFR | | | | 0.1238 (0.9015) | -10.4080 (0.0000) | -13.7249 (0.0000) |
| LSTM | | | | | -10.3531 (0.0000) | -13.2739 (0.0000) |
| SVR | | | | | | 1.7813 (0.0760) |

**Table 10** DM test results of Dataset 3

| Tested models | Reference model | | | | | |
|---|---|---|---|---|---|---|
| | VMD-LSTM | VMD-SVR | VMD-RFR | LSTM | SVR | RFR |
| VMD-CNN-LSTM | -1.6334 (0.1035) | -4.9189 (0.0000) | -5.9252 (0.0000) | -6.2958 (0.0000) | -6.0660 (0.0000) | -5.8727 (0.0000) |
| VMD-LSTM | | -4.2855 (0.0000) | -5.4927 (0.0000) | -6.1236 (0.0000) | -5.8882 (0.0000) | -5.6842 (0.0000) |
| VMD-SVR | | | -5.8110 (0.0000) | -5.9545 (0.0000) | -5.7481 (0.0000) | -5.4846 (0.0000) |
| VMD-RFR | | | | -5.1023 (0.0000) | -5.1320 (0.0000) | -4.4946 (0.0000) |
| LSTM | | | | | 0.0024 (0.9981) | 1.4745 (0.1414) |
| SVR | | | | | | 1.0965 (0.2738) |

**Table 11** DM test results of Dataset 4

| Tested models | Reference model | | | | | |
|---|---|---|---|---|---|---|
| | VMD-LSTM | VMD-SVR | VMD-RFR | LSTM | SVR | RFR |
| VMD-CNN-LSTM | -1.8188 (0.0699) | -5.1509 (0.0000) | -5.5344 (0.0000) | -6.6981 (0.0000) | -6.4719 (0.0000) | -6.8203 (0.0000) |
| VMD-LSTM | | -5.0186 (0.0000) | -5.4404 (0.0000) | -6.6650 (0.0000) | -6.4412 (0.0000) | -6.7868 (0.0000) |
| VMD-SVR | | | -4.0144 (0.0000) | -6.1253 (0.0000) | -6.0701 (0.0000) | -6.4436 (0.0000) |
| VMD-RFR | | | | -6.1135 (0.0000) | -5.9969 (0.0000) | -6.3681 (0.0000) |
| LSTM | | | | | -2.2026 (0.0284) | -0.7471 (0.4556) |
| SVR | | | | | | 1.3154 (0.1894) |

In summary, the above results present the following implications. Firstly, embedding the single forecasting step and ensemble step into a unified deep learning approach is effective to improve the forecasting performance. Secondly, the decomposition-reconstruction-ensemble framework may be better than the decomposition-ensemble framework for time series forecasting problems. Thirdly, the integrated approaches significantly show better performance than the single models. Finally, the LSTM based approaches always show better performance than other forecasting methods based approaches.

## 5. Conclusions

In this paper, we propose a new VMD based deep learning approach for time series forecasting. In the proposed VMD-CNN-LSTM approach, VMD is adopted to decompose the original time series into several sub-signals, CNN is applied to learn the reconstruction patterns from the decomposition sub-signals to obtain several reconstructed sub-signals, and LSTM is employed to forecast time series with the decomposed sub-signals and the reconstructed sub-signals as inputs. The proposed VMD-CNN-LSTM approach is based on the idea of decomposition-reconstruction-ensemble framework, and the innovation is that the reconstruction, single forecasting, and ensemble steps in the framework are embedded in a unified deep learning approach.

Two crude oil price datasets and two wind speed datasets are collected as the experimental time series datasets to verify the performance of the proposed approach. The empirical results show that the proposed VMD-CNN-LSTM approach outperforms consistently the benchmark approaches in terms of forecasting accuracy, which demonstrates the effectiveness of the proposed approach. It also indicates that it is an effective way to significantly improve the forecasting performance by embedding the single forecasting step and ensemble step into a unified deep learning approach. In addition, the reconstructed sub-signals obtained by CNN is important for the proposed approach to further improve the forecasting performance.

Furthermore, since the proposed VMD-CNN-LSTM shows better performance, it may implicate that decomposition-reconstruction-ensemble framework could be a better choice for other time series forecasting problems. Therefore, the decomposition-reconstruction-ensemble framework based approaches are a promising choice for time series forecasting problem, and deserve more research attentions in the future. In addition, the proposed VMD-CNN-LSTM approach can be applied to solve other complex and difficult time series forecasting problems, except for the time series introduced in this paper for empirical analysis.

However, we only consider a simple reconstruction structure in the proposed approach in this paper, while more complex structures can be investigated in future research. Furthermore, other neural networks can also be applied for single forecasting and ensemble steps instead of a simple LSTM, which is adopted in this paper.

**Conflict of Interests**

The authors declare that there is no conflict of interests regarding the publication of this paper.